\documentclass[letterpaper, 10 pt, conference]{ieeeconf}  % Comment this line out if you need a4paper

\IEEEoverridecommandlockouts                              % This command is only needed if 
\usepackage{times} % assumes new font selection scheme installed
\usepackage{amsmath} % assumes amsmath package installed
\usepackage{amssymb}  % assumes amsmath package installed
\usepackage{graphicx}
\usepackage{subcaption}
\usepackage{color}
\usepackage{lipsum}
\usepackage{cases}
\usepackage{algorithm}
\usepackage{algorithmic}
\usepackage{booktabs}
\usepackage{multirow}

\title{\LARGE \bf
Safety Correction from Baseline: Towards the Risk-aware Policy in Robotics via Dual-agent Reinforcement Learning
}

\author{Linrui Zhang$^{*}$, Zichen Yan$^{*}$, Li Shen, Shoujie Li, Xueqian Wang$^\dag$ and Dacheng Tao% <-this % stops a space
\thanks{*These authors contributed eqaully to this work.}% <-this % stops a space
\thanks{$\dag$Corresponding author, email: wang.xq@sz.tsinghua.edu.cn}% <-this % stops a space
\thanks{This work is supported by the Major Science and Technology Innovation 2030 “Brain Science and Brain-like Research” key project (No. 2021ZD0201405), Shenzhen Philosophy and Social Science Foundation (Grant No. SZ2021B005).}
\thanks{Linrui Zhang, Zichen Yan, Shoujie Li and Xueqian Wang are with the Center for Intelligent Control and Telescience, Tsinghua Shenzhen International Graduate School, 518055, Shenzhen, China.}%
\thanks{Li Shen and Dacheng Tao are with the JD Explore Academy, 100176, Beijing, China.}%
}

\begin{document}

\maketitle
\thispagestyle{empty}
\pagestyle{empty}

%%%%%%%%%%%%%%%%%%%%%%%%%%%%%%%%%%%%%%%%%%%%%%%%%%%%%%%%%%%%%%%%%%%%%%%%%%%%%%%%
\begin{abstract}
Learning a risk-aware policy is essential but rather challenging in unstructured robotic tasks. Safe reinforcement learning methods open up new possibilities to tackle this problem. However, the conservative policy updates make it intractable to achieve sufficient exploration and desirable performance in complex, sample-expensive environments. % motivation
In this paper, we propose a dual-agent safe reinforcement learning strategy consisting of a baseline and a safe agent.
Such a decoupled framework enables high flexibility, data efficiency and risk-awareness for RL-based control.  % our approach
Concretely, the baseline agent is responsible for maximizing rewards under standard RL settings. Thus, it is compatible with off-the-shelf training techniques of unconstrained optimization, exploration and exploitation. On the other hand, the safe agent mimics the baseline agent for policy improvement and learns to fulfill safety constraints via off-policy RL tuning. In contrast to training from scratch, safe policy correction requires significantly fewer interactions to obtain a near-optimal policy. 
The dual policies can be optimized synchronously via a shared replay buffer, or leveraging the pre-trained model or the non-learning-based controller as a fixed baseline agent. % detailed solution and benefits
Experimental results show that our approach can learn feasible skills without prior knowledge as well as deriving risk-averse counterparts from pre-trained unsafe policies. The proposed method outperforms the state-of-the-art safe RL algorithms on difficult robot locomotion and manipulation tasks with respect to both safety constraint satisfaction and sample efficiency.
\end{abstract}

%%%%%%%%%%%%%%%%%%%%%%%%%%%%%%%%%%%%%%%%%%%%%%%%%%%%%%%%%%%%%%%%%%%%%%%%%%%%%%%%
\section{INTRODUCTION}
\label{sec:1}
Learning-based methods have achieved significant success in many long-standing, difficult robotic tasks~\cite{francis2020long,rajeswaran2017learning,shin2019obstacle}. These advances also raise concerns about the safety of autonomous agents in practical applications: It makes sense to maximize total rewards in the context of a specific task, but is very likely to cause unexpected side-effects to the surroundings~\cite{amodei2016concrete}. 
% Imagine that a desktop cleaning robot has learned to move waste to the garbage bin, but it would not deliberately avoid knocking over a vase around it. Hence, learning a risk-aware policy means keeping away "vases", namely the unstructured risks in the environment.
In this paper, we focus on the topic of learning ``safety'' in order to obtain a stationary policy that can finish the original task and deliberately avoid the risks in the environment, which is distinguished from learning ``safely''. This problem used to be solved with robust model predictive control~\cite{hewing2020learning}, Gaussian process regression~\cite{sui2015safe}, etc. Safe reinforcement learning (Safe RL) opens up new possibilities for risk-aware policy optimization and continues to gain traction in recent literature~\cite{brunke2021safe}. 

Safe RL aims to maximize cumulative rewards under the precondition that all the given safety constraints are satisfied. However, it is cumbersome to solve such a constrained sequential decision-making problem in a large parametric space, because most of the existing algorithms are sample inefficient due to the on-policy episodic cost evaluation~\cite{ray2019benchmarking} and conservative policy updates~\cite{DBLP:conf/icml/AchiamHTA17}. Even worse, many off-the-shelf training techniques in standard RL are less studied in this scope and do not necessarily work when considering the safety constraints.
Consequently, directly applying safe RL to robotic tasks is intractable and usually stuck in a dilemma: The agent needs adequate experiences by trial-and-error to master complex skills, but the safe RL algorithm often limits its exploration within safety-proven behaviors, and the poor data efficiency also makes it impractical on risky, sample-expensive robot learning scenarios.

\begin{figure}
      \centering
      \subcaptionbox{Rollout of Baseline Policy\label{fig:directpolicy}}
        {\includegraphics[width=0.475\linewidth]{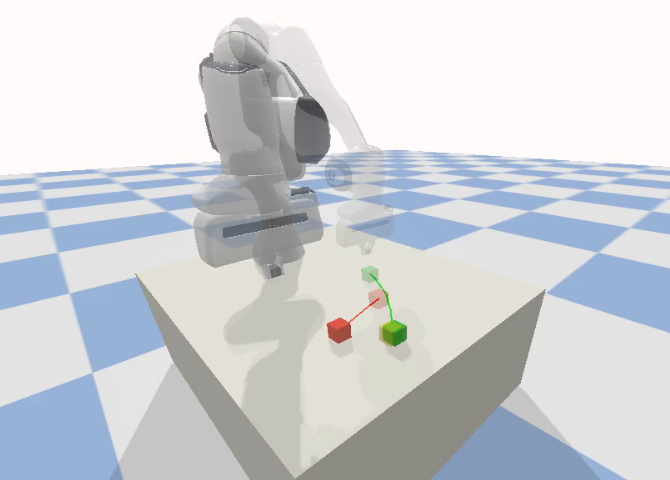}}
      \subcaptionbox{Rollout of Safe Policy\label{fig:safepolicy}}
        {\includegraphics[width=0.475\linewidth]{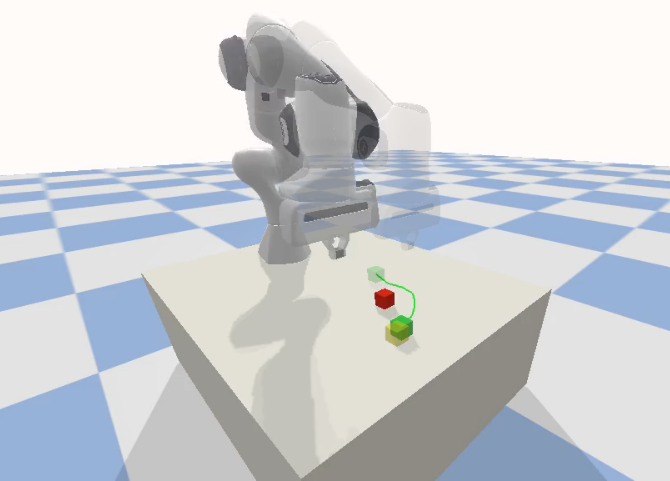}}
      \caption{Illustration of the risk-aware policy correction. The robotic arm is required to push the green cube to the destination in the yellow shadow. The baseline policy will move the object along the shortest path, regardless of the obstacle in its way. The safe policy will correct the trajectory slightly to a curve to avoid potential risks instead.}
      \label{fig:intro}
      \vspace{-0.5cm}
\end{figure}

As the example of safe manipulation in Fig.~\ref{fig:intro}, we notice that in most robot learning tasks, the safe policy (considering safety constraints) is similar to the baseline policy (ignoring safety constraints) on the general trend, but with modified actions or trajectories in a few dangerous situations.
Motivated by that, we propose a dual-agent algorithm as the alternative for previous safe RL to tackle the above issues in robotic tasks. Concretely, one agent (i.e., the baseline agent) explores the environment and only aims to maximize cumulative rewards. The other (i.e., the safe agent) learns to satisfy given constraints while mimicking the former via online imitation learning to obtain basic skills quickly.

From the perspective of optimization, separating the optimization objective and the constraint can degrade the difficulty of obtaining a feasible solution. We summarize the strengths of the proposed decoupled framework as four-fold: (1) The baseline policy learning is an independent and unconstrained RL task, where most of the training techniques and exploration strategies are available. (2) The difficulty of finding a near-optimal safe policy is reduced due to the online behavior cloning at the early stage and the synchronous constrained RL tuning. (3) The baseline policy guides the safe policy correction and can be in any form. Thus, it is especially pragmatic when we already have pre-trained models or non-learning-based controllers without specific safety considerations, which are common and well-studied in many standard robotic tasks. (4) When it comes to new risky environments, we can execute the safe policy correction with the baseline policy frozen. This significantly increases the sample efficiency compared to training from scratch.

To demonstrate the efficacy of our approach, we design and conduct experiments on different safe robot learning tasks. For the quadruped robot locomotion task, the baseline policy learning and safe policy correction are performed simultaneously without any prior knowledge. For the robotic arm manipulation task, the safe policy is corrected based on a fixed baseline policy which is pre-trained with Hindsight Experience Replay (HER)~\cite{andrychowicz2017hindsight} technique. The empirical results show that our approach can find a near-optimal policy through very limited interactions and outperforms the state-of-the-art safe RL algorithms with respect to safety constraint satisfaction and reward improvement.

\section{RELATED WORK}
\subsection{Safe Reinforcement Learning}
The safety problem in reinforcement learning has become a research hot spot in recent years~\cite{DBLP:conf/ijcai/LiuHL21}. In most works, it is regarded as a constrained RL problem, where the agent receives additional cost signals and learns to satisfy the cost constraints. A popular way is leveraging Lagrangian duality for hard constraints~\cite{9029423}, but it is rather data inefficient and sensitive to Lagrangian multipliers. As an alternative, Constrained Policy Optimization (CPO) \cite{DBLP:conf/icml/AchiamHTA17} directly searches the feasible policy in the trust-region and guarantees a monotonic performance improvement while satisfying constraints by solving an approximated quadratic optimization problem. Unfortunately, this method suffers from approximation errors and heavy computational burdens for Hessian Matrix inversion, making it incompetent for complex tasks. Another approach is to correct the action at each step by projecting it onto a feasible space. It can be implemented by adding a safety layer \cite{DBLP:journals/corr/abs-1801-08757} or solving a quadratic programming problem \cite{DBLP:conf/iros/HirshbergVK20}. In contrast to training safe policies from scratch, it is much easier to obtain an unconstrained baseline policy and locally modifies it for safety.

\subsection{Reinforcement Learning from Demonstration}
Since we use the baseline policy as guidance, the study of Reinforcement Learning from Demonstration (RLfD) is also related to our work. Previous methods \cite{DBLP:conf/aaai/HesterVPLSPHQSO18,DBLP:journals/corr/VecerikHSWPPHRL17,DBLP:conf/icml/KangJF18} try to combine behavioral cloning (BC) with RL to speed up training and improve the exploration efficiency. Extensive studies \cite{DBLP:conf/aaai/JingMHSY0020,DBLP:conf/iclr/GaoXLYLD18,DBLP:conf/atal/GoecksGLVW20} 
show that even imperfect demonstrations are allowed to guide the policy improvement. The key principle of those methods is to find an optimal policy that maximizes the return while mimicking the expert strategy, where a combined loss function of RL and supervised learning is widely used for training. However, such a setting is sensitive to the initialization of the weight coefficients and the learning rate.
Furthermore, \cite{DBLP:conf/ijcai/Wang0ZCZ21} considers safety issues in RLfD to avoid undesirable behaviors, where Lagrangian relaxation is applied to solve the constrained optimization problem. 

\subsection{Decoupled Policy Learning}

Decoupled reinforcement learning (DeRL) is a very new concept and formally raised in~\cite{schafer2021decoupling}. The main purpose of their decoupled framework is to solve the dilemma of exploration and exploitation in meta reinforcement learning. A similar idea can be referred to~\cite{whitney2021decoupled}, which demonstrates decoupling exploration with policy learning enables a several-fold improvement in data efficiency on sparse reward environments. To the best of our knowledge, our approach is the first algorithm to apply decoupled policy learning to the safety of autonomous systems.
\section{PROBLEM FORMULATION}
In this work, we consider the risk-aware policy optimization in an constrained Markov Decision Process~\cite{altman1999constrained}, represented by a tuple $(\mathcal{S,A,R,P,C})$. $\mathcal{S}$ and $\mathcal{A}$ denote the state space and the action space, respectively. $\mathcal{R}:  \mathcal{S} \times  \mathcal{A} \times  \mathcal{S}  \mapsto \mathbb{R}$ is the reward function and $\mathcal{P}: \mathcal{S} \times  \mathcal{A} \times  \mathcal{S} \mapsto [0,1]$ is the transition  probability  function to describe the dynamics of the environment. $\mathcal{C} : \mathcal{S} \times \mathcal{A}\times \mathcal{S}\mapsto [0,+\infty]$ is a cost function, that reflects the violation of safety requirements.

The stochastic policy $\pi(\cdot | s)$ maps the given state $s$ to a probability distribution over action space (using Dirac delta distribution for the deterministic policy). The goal of the agent is to maximize the expected discounted return
$J_R(\pi) = \mathop{\mathbb{E}}_{\tau\sim \pi}\big [ \sum^T_{t=0}\gamma^t r_t\big ]$. Here $\tau=\{(s_t,a_t,r_t)\}^T_{t\ge0}$ is a sampled trajectory and the discounted factor $\gamma\in (0,1]$ guarantees the geometric series vanishes when the time horizon $T$ goes towards infinity. The value function is defined as $V_R^\pi (s) = \mathop{\mathbb{E}}_{\tau\sim \pi}\big [ \sum^T_{k=t}\gamma^k r_k | s_t = s\big]$, and the action-value function is defined as $Q_R^\pi (s,a) = \mathop{\mathbb{E}}_{\tau\sim \pi}\big [ \sum^T_{k=t}\gamma^k r_k | s_t = s,a_t =a \big]$. Then the optimization problem of RL can be formulated as:
\begin{equation}
\mathop{\max}_{\pi} \mathop{\mathbb{E}}_{s \sim \tau}\big [V_R^\pi (s) \big]
\label{maxq}
\end{equation}

As is discussed in Section~\ref{sec:1}, solving problem~\eqref{maxq} often leads to a unsafe policy with the potential risk to the environment or the agent itself.
Therefore, we define an safety indicator $I$ at each time-step:
\begin{equation}
\label{safeI}
I(s_t,a_t)=\left\{
\begin{aligned}
&\  1 \quad \mathcal{C}(s_t,a_t,s_{t+1}) = 0 \\
&\  0 \quad \mathcal{C}(s_t,a_t,s_{t+1}) > 0\\
\end{aligned}
\right.
\end{equation}
Then we give the following definition for state-wise safety.

%\begin{definition}
\textbf{Definition: }The policy $\pi$ is $(1-\delta)$-safe if for each time-step $t$ we have
\begin{equation}
\mathbb{E}[I(s_t,a_t)] \ge 1-\delta.
\label{EI}
\end{equation}
%\end{definition}

Considering the impact of sequential decision-making, we define $V_I^\pi (s) = \mathop{\mathbb{E}}_{\tau\sim \pi}\big [ \sum^T_{k=t}\bar\gamma^k I_k | s_t = s\big]$, and then reformulate~\eqref{EI} in the context of RL:
\begin{equation}
V_I^\pi (s)  =  \sum_{k=t}^T \bar\gamma^k \mathbb{E}\big[I_k|s_t=s\big] \ge
(1-\delta) \frac{1-\bar\gamma^{T-k}}{1-\bar\gamma}
\label{EVI1}
\end{equation}

Thus, we derive the expectation of $V_I^\pi (s)$ sampled from experience replay buffer as:
%\begin{equation}
\begin{align}
\mathop{\mathbb{E}}_{s \sim \tau}\big [V_I^\pi (s) \big] &\ge \frac{1}{T} \sum_{k=0}^{T-1} (1-\delta) \frac{1-\bar\gamma^{T-k}}{1-\bar\gamma} \label{tmp}\\
&= (1-\delta)\frac{T\cdot(1-\bar\gamma)-\bar\gamma(1-\bar\gamma^T)}{T\cdot(1-\bar\gamma)^2}\label{EVI}
\end{align}
%\end{equation}
If $T \rightarrow +\infty$, we have the following approximation:
\begin{equation}
    \mathop{\mathbb{E}}_{s \sim \tau}\big [V_I^\pi (s) \big] \ge (1-\delta)/(1-\bar\gamma)
\end{equation}

Hence, the safe robot learning problem is given by:
\begin{equation}
\begin{aligned} \label{P}
\max_{\pi}& \mathop{\mathbb{E}}_{s \sim \tau}\big [V_R^\pi (s) \big]\\
%&\begin{array}{r@{\quad}r@{}l@{\quad}l}
\mathrm{s.t.} &\mathop{\mathbb{E}}_{s \sim \tau}\big [V_I^\pi (s) \big] \ge \Gamma\\
%\end{array}
\end{aligned}
\end{equation}
$\Gamma$ is a consistent associated with $(1-\delta)$-safe requirement:
\begin{equation}
\Gamma = (1-\delta)\frac{T\cdot(1-\bar\gamma)-\bar\gamma(1-\bar\gamma^T)}{T\cdot(1-\bar\gamma)^2}
\label{Gamma}
\end{equation}

\section{METHODS}
\label{sec:method}

\subsection{Dual-agent Framework for Safe Robot Learning}
Directly solving problem~\eqref{P} is feasible theoretically but intractable practically in safe robot learning. Because it often struggles with the dilemma of exploration and exploitation: Mastering a difficult task requires adequate exploration in unknown environments, but the hard constraint restricts output actions in a small, safety-proven region. Besides, the conservative policy updates limit the data efficiency, making it impractical on complex and sample-expensive scenarios.

\begin{figure}
      \centering
        \includegraphics[width=0.95\linewidth]{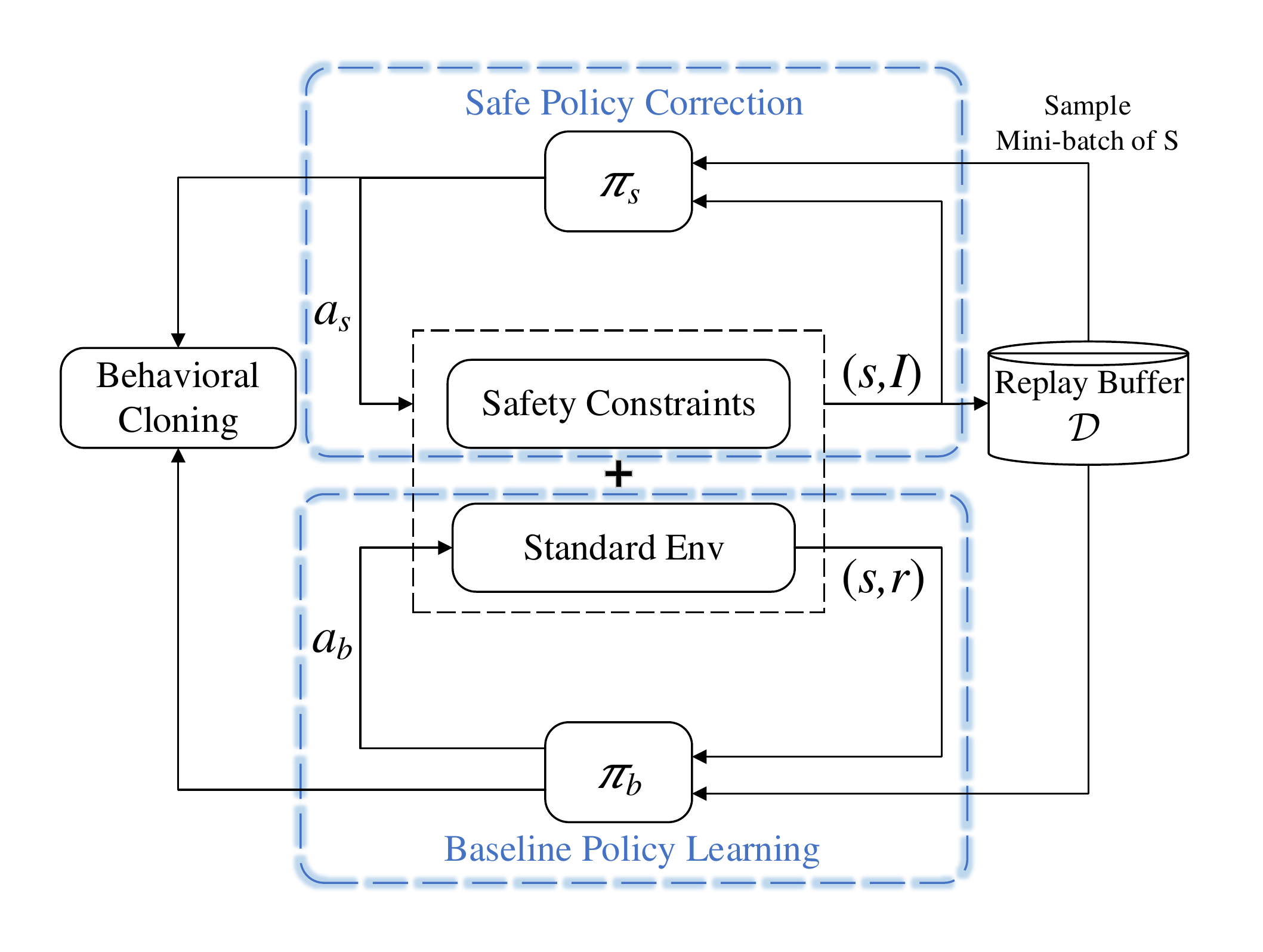}
      \caption{Dual-agent framework for safe robot learning. The baseline policy learning is the standard RL whose objective is the maximization of cumulative rewards. The safe policy correction learns safe behaviors from safety indicator $I$ and imitates the baseline policy via online behavioral cloning.}
      \label{fig:framework}
\end{figure}

As we argued in Section~\ref{sec:1}, in most robotic tasks, the risk-aware policy is usually similar to the unconstrained one in general, but with modified outputs to avoid dangerous situations. As is illustrated in Fig.~\ref{fig:framework}, we decouple the safe robot learning problem~\eqref{P} into two separated agents: One agent (in the dashed box below) aims to learn the baseline policy $\pi_B$ that maximizes cumulative rewards:  
\begin{equation}\label{bpl}
    \max_{\pi_B} \mathop{\mathbb{E}}_{s \sim \tau}\big [V_R^{\pi_B} (s) \big],
\end{equation}
the other agent (in the dashed box above) aims to learn the risk-aware policy $\pi_S$ from interactions while getting close to the baseline policy as possible:
\begin{equation}
\begin{aligned} \label{spc}
\min_{\pi_S}& \ \ \mathrm{D}(\pi_S || \pi_B)\\
%&\begin{array}{r@{\quad}r@{}l@{\quad}l}
\mathrm{s.t.} &\mathop{\mathbb{E}}_{s \sim \tau}\big [V_I^{\pi_S} (s) \big] \ge \Gamma \\
%\end{array}
\end{aligned}
\end{equation}

Notably, we can take advantage of pre-trained models or non-learning-based controllers as guidance. Nevertheless, we will focus on the case that two agents are trained simultaneously from scratch in this section.

\subsection{Baseline Policy Learning}
 
Baseline policy learning, i.e., problem~\eqref{bpl}, is a basic RL task that can be solved with a bunch of effective techniques in optimization, exploration and exploitation. 

Considering a parametric deterministic policy $\pi_B(s;\theta)$, problem~\eqref{bpl} is equivalent to:
\begin{equation}\label{bpl2}
    \max_{\theta} \mathop{\mathbb{E}}_{s \sim \mathcal{D}}\big [Q_R^{\pi_B} \big(s,\pi_B(s;\theta)\big) \big],
\end{equation}
whose parameters can be updated by the deterministic policy gradient (DPG)~\cite{silver2014deterministic}:
\begin{equation}\label{bpl3}
    \theta\leftarrow \theta + \eta_\theta \nabla_\theta \pi_B(s;\theta)\big [ \nabla_a Q_R^{\pi_B} (s,a)\big]_{a=\pi_B(s;\theta)}
\end{equation}

\subsection{Safe Policy Correction}

% \begin{proposition}
% \zlr{the distance between two policies ...}
% \end{proposition}

% \begin{proof}
% See the Appendix.
% \end{proof}
Safe policy correction, i.e. problem~\eqref{spc}, is a constrained optimization problem. Considering a parametric deterministic policy $\pi_S(s;\phi)$ and the $L^2$-norm distance, problem~\eqref{spc} can be reformulated as:
\begin{equation}
\begin{aligned} \label{spc2}
\min_{\phi}& \mathop{\mathbb{E}}_{s \sim \mathcal{D}} ||\pi_S(s;\phi) - \pi_B(s;\theta)||^2_2\\
%&\begin{array}{r@{\quad}r@{}l@{\quad}l}
\mathrm{s.t.} &\mathop{\mathbb{E}}_{s \sim \mathcal{D}}\big [Q_I^{\pi_S} \big(s,\pi_S(s;\phi)\big) \big] \ge \Gamma \\
%\end{array}
\end{aligned}
\end{equation}

Similar to the strong Lagrangian duality in~\cite{9029423}, the primal problem~\eqref{spc2} can be solved through its dual problem:
\begin{equation}
  \max_{\lambda\ge 0} \min_{\phi} \bar D(\phi) + \lambda (\Gamma - \bar Q(\phi) ),  
\end{equation}
where $\bar D(\phi) = \mathop{\mathbb{E}}_{s \sim \mathcal{D}} ||\pi_S(s;\phi) - \pi_B(s;\theta)||^2_2$ and $\bar Q(\phi) = \mathop{\mathbb{E}}_{s \sim \mathcal{D}}\big [Q_I^{\pi_S} \big(s,\pi_S(s;\phi)\big)\big ]$.

Stochastic primal-dual optimization~\cite{luenberger1984linear} is applied here to update primal and dual variables alternatively:

\begin{numcases}{}
&$\phi \leftarrow \phi - \eta_\phi \nabla_\phi \big[ \bar D(\phi) -\lambda \bar Q(\phi) ) \big]$ \label{phi}\\
&$\lambda \leftarrow \big[ \lambda + \eta_\lambda  (\Gamma - \bar Q(\phi) ) \big]^+$ \label{lam}
\end{numcases}

\begin{algorithm}[h]
\caption{Dual-agent Risk-aware Policy Learning.}
\begin{algorithmic}[1]
\small{
%\REQUIRE Safe critic $Q_{\xi_1},Q_{\xi_2}$ and target policy $\pi_\phi$.
\STATE Initialize safe critics $Q_{\xi_1},Q_{\xi_2}$ and the risk-aware policy $\pi_\phi$.
\STATE Initialize target networks ${\xi'_1}\leftarrow{\xi_1},{\xi'_2}\leftarrow{\xi_2},{\phi'}\leftarrow{\phi}$.
\STATE Initialize the replay buffer $\mathcal{D}$, $\lambda=0$ and $\Gamma$ as in~\eqref{EVI}.
\IF{the baseline policy $\pi_\theta$ is learnable} 
\STATE Initialize critic $Q_{\omega_1},Q_{\omega_2}$ and the baseline policy $\pi_\theta$.
\STATE Initialize target networks ${\omega'_1}\leftarrow{\omega_1},{\omega'_2}\leftarrow{\omega_2},{\theta'}\leftarrow{\theta}$.
\ELSE
\STATE Load the fixed baseline policy $\pi_\theta$.
\ENDIF
\FOR{$t = 1$ \textbf{to} $T$}
\IF {$\upsilon \sim \mathcal{U}(0,1) < 0.5$}
\STATE Select action $a\sim \pi_\theta(s) + \epsilon,\ \ \epsilon\sim\mathcal{N}(0,\sigma)$.
\ELSE
\STATE Select action $a\sim \pi_\phi(s) + \epsilon,\ \ \epsilon\sim\mathcal{N}(0,\sigma)$.
\ENDIF
\STATE Calculate safe indicator $I$ as in~\eqref{safeI}.
\STATE Store transition $(s,a,r,I,s')$ in the replay buffer $\mathcal{D}$.
\STATE Sample mini-batch of N transitions from $\mathcal{D}$.
\STATE $//$ Baseline Policy Learning;
\IF{the baseline policy $\pi_\theta$ is learnable}
\STATE $y\leftarrow r + \gamma \min_{i=1,2}Q_{\omega'_i}(s',\pi_{\theta'}(s'))$.
\STATE Update $\omega_i \leftarrow\arg\min_{\omega_i}N^{-1}\sum(y-Q_{\omega_i}(s,a))^2$.
\STATE Update $\theta$ as in~\eqref{bpl3}.
\STATE Update target networks with Polyak Averaging.
\ENDIF
\STATE $//$ Safe Policy Correction;
\STATE $\bar y\leftarrow I + \bar\gamma \min_{i=1,2}Q_{\xi'_i}(s',\pi_{\phi'}(s'))$.
\STATE Update $\xi_i \leftarrow\arg\min_{\xi_i}N^{-1}\sum(\bar y-Q_{\xi_i}(s,a))^2$.
\STATE Calculate $\bar D(\phi) = \mathop{\mathbb{E}}_{s \sim \mathcal{D}} ||\pi_S(s;\phi) - \pi_B(s;\theta)||^2_2$.
\STATE Update $\phi$ as in~\eqref{phi}.
\STATE Evaluate $\bar Q(\phi) = \mathop{\mathbb{E}}_{s \sim \mathcal{D}}\big [Q_I^{\pi_S} \big(s,\pi_S(s;\phi)\big)\big ]$.
\STATE Update $\lambda$ as in~\eqref{lam}.
\STATE Update target networks with Polyak Averaging.
\ENDFOR
}
\label{algo}
\end{algorithmic}
\end{algorithm}

Technically, the timescales of primal variable updates are required to be faster than those of Lagrange multipliers (i.e., $\eta_\phi \gg \eta_\lambda$)~\cite{tessler2018reward} to replace the computationally prohibitive minimization of primal variables by a single gradient descent step~\eqref{phi}.

\subsection{Further Discussion and Detailed Algorithm}
We provide further discussions on several useful tricks and insights for better policy improvement in practice. The detailed algorithm is summarized in Algorithm~\ref{algo}. 
\subsubsection{Initialization}
Primal-dual optimization is sensitive to the initialization and learning rate of Lagrange multipliers. In our implementation, we set $\lambda = 0$ at the beginning and let $\eta_\lambda = 0.001$, which enables the agent to learn skills from the baseline policy rapidly and correct dangerous behaviors progressively through interactions. 

\subsubsection{Exploration}
The decoupled framework enables each agent to have an independent reinforcement learning objective, which facilitates the exploration naturally. We suggest selecting actions randomly and alternatively from the two policies, thus each is likely to see more novel and valuable state-action pairs. Moreover, other exploration strategies can be applied directly to the baseline policy learning as in~\cite{yang2021exploration}. 

\subsubsection{Over-estimation}
The over-estimation due to off-policy Q-learning is catastrophic to $Q^{\pi_S}_I$ evaluation. To address this issue, the same technique of double Q-learning in~\cite{fujimoto2018addressing} is applied, and we suggest a small $\bar\gamma$ which is myopic for future safety.

\subsubsection{Optimality}
Decoupling baseline policy learning and safe policy correction isn't strictly equivalent to the safe RL problem~\eqref{P}. Thus, our approach only finds a near-optimal policy compared to the primal formulation. Nevertheless, the desirable performance with high sample efficiency usually outweighs the theoretical optimality for most safe robot learning tasks.

\begin{figure*}
      \centering
      \subcaptionbox*{$\pi_b:t=0$}
        {\includegraphics[width=0.15\linewidth,trim=0 50 0 0,clip]{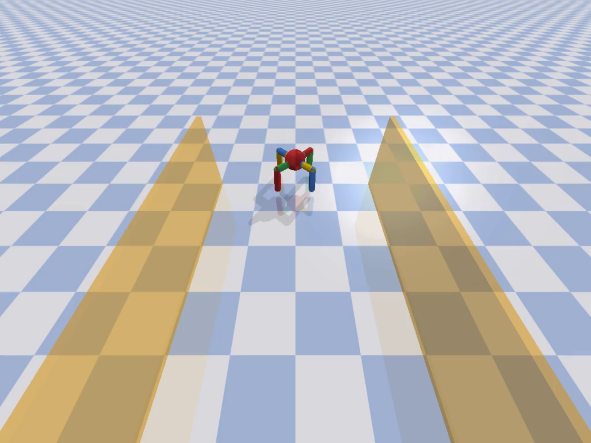}}
        \subcaptionbox*{$\pi_b:t=100$}
        {\includegraphics[width=0.15\linewidth,trim=0 50 0 0,clip]{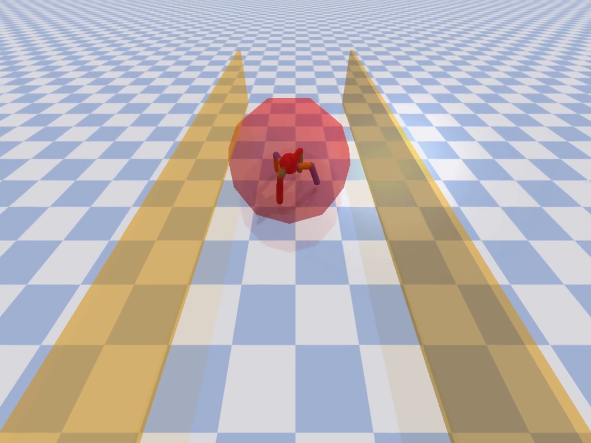}}
          \subcaptionbox*{$\pi_b:t=200$}
        {\includegraphics[width=0.15\linewidth,trim=0 50 0 0,clip]{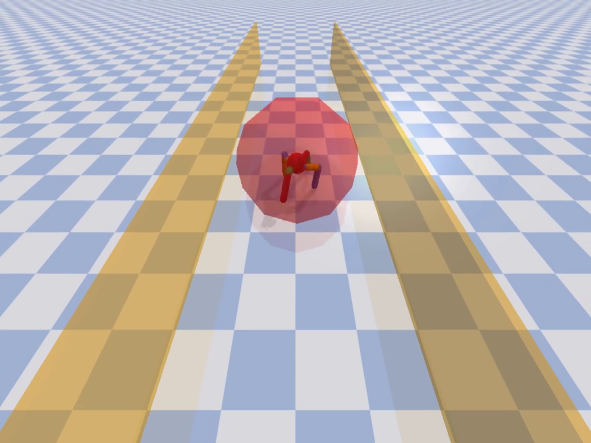}}
        \subcaptionbox*{$\pi_b:t=300$}
        {\includegraphics[width=0.15\linewidth,trim=0 50 0 0,clip]{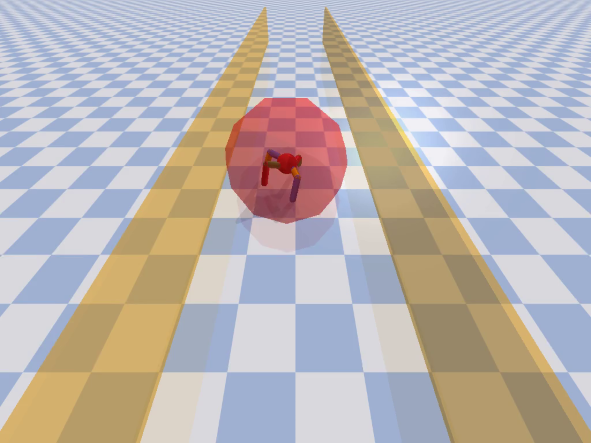}}
              \subcaptionbox*{$\pi_b:t=400$}
        {\includegraphics[width=0.15\linewidth,trim=0 50 0 0,clip]{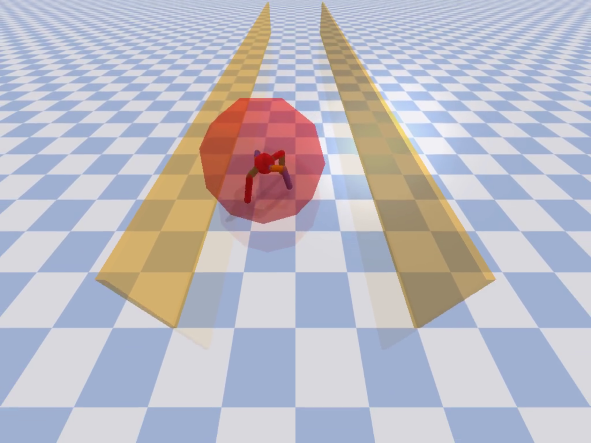}}
        \subcaptionbox*{$\pi_b:t=500$}
        {\includegraphics[width=0.15\linewidth,trim=0 50 0 0,clip]{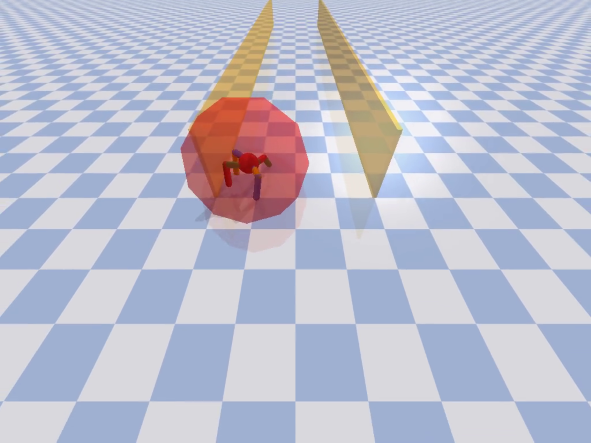}}
      \\  
    \subcaptionbox*{$\pi_s:t=0$}
        {\includegraphics[width=0.15\linewidth,trim=0 50 0 0,clip]{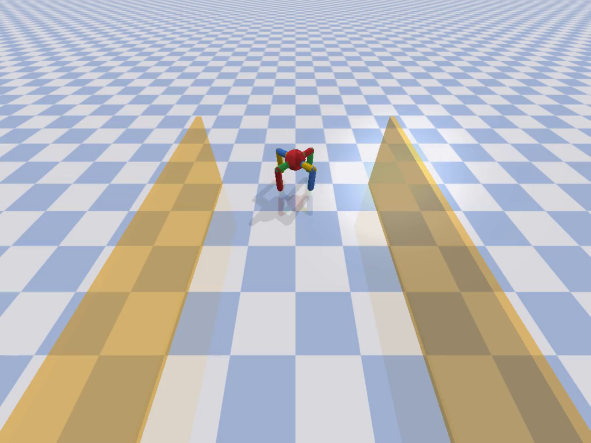}}
        \subcaptionbox*{$\pi_s:t=100$}
        {\includegraphics[width=0.15\linewidth,trim=0 50 0 0,clip]{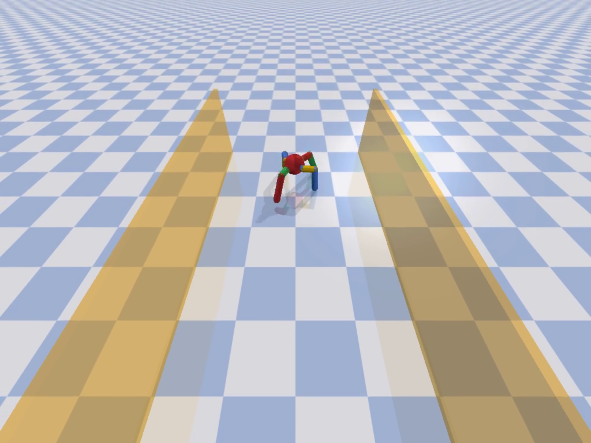}}
          \subcaptionbox*{$\pi_s:t=200$}
        {\includegraphics[width=0.15\linewidth,trim=0 50 0 0,clip]{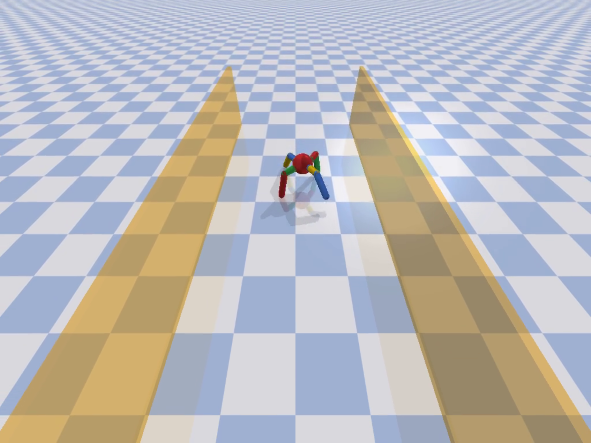}}
        \subcaptionbox*{$\pi_s:t=300$}
        {\includegraphics[width=0.15\linewidth,trim=0 50 0 0,clip]{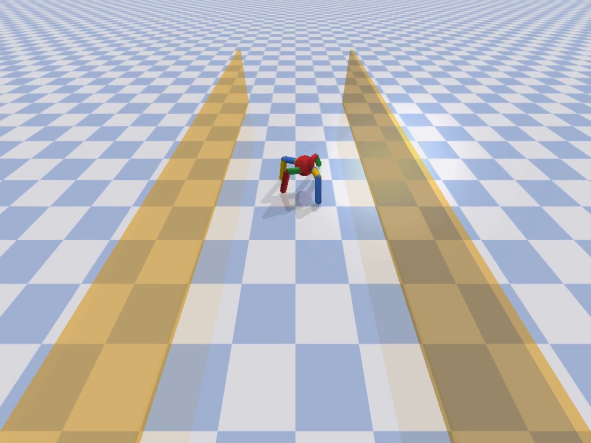}}
              \subcaptionbox*{$\pi_s:t=400$}
        {\includegraphics[width=0.15\linewidth,trim=0 50 0 0,clip]{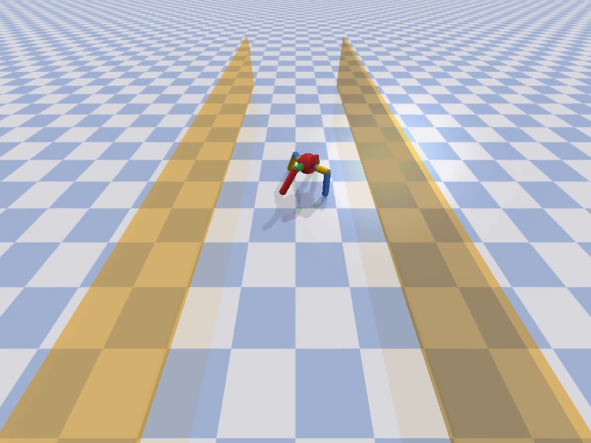}}
        \subcaptionbox*{$\pi_s:t=500$}
        {\includegraphics[width=0.15\linewidth,trim=0 50 0 0,clip]{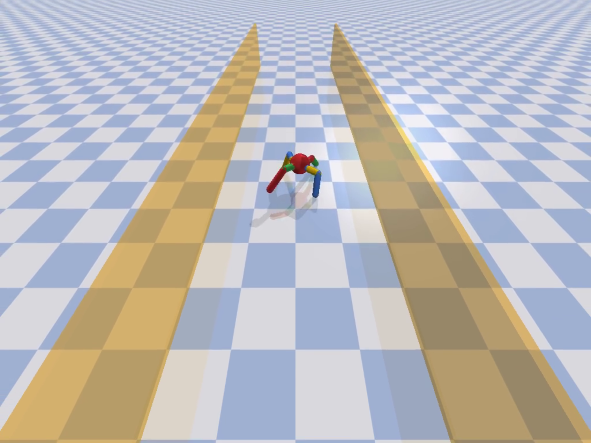}}
      \caption{Quadruped robot locomotion task. The agent is rewarded for running through an avenue~\cite{chow2019lyapunov}, but violates the safety requirement (marked with a red ball) if it oversteps the non-physical boundaries or exceed the velocity threshold. Rollouts of the baseline policy $\pi_B$ and the safe policy $\pi_S$ are shown on the top line and bottom line, respectively.}
      \label{fig:ant}
\end{figure*}
\begin{figure*}
      \centering
      \subcaptionbox*{$\pi_b:t=0$}
        {\includegraphics[width=0.15\linewidth]{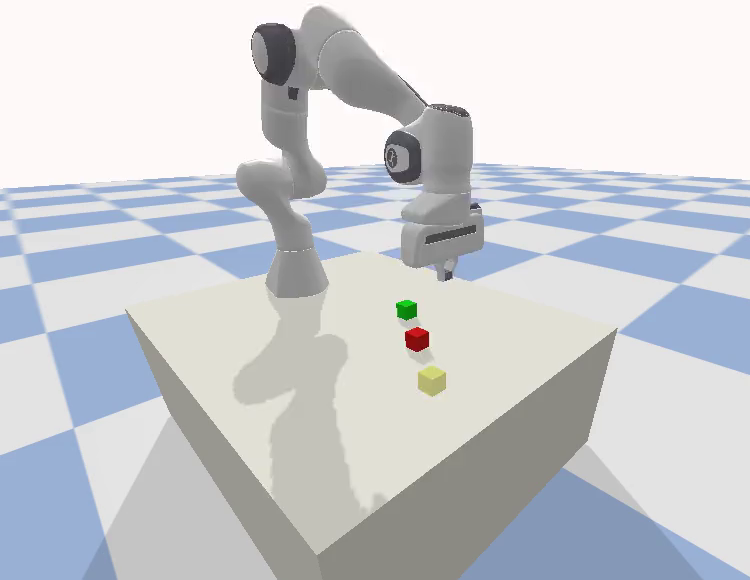}}
        \subcaptionbox*{$\pi_b:t=10$}
        {\includegraphics[width=0.15\linewidth]{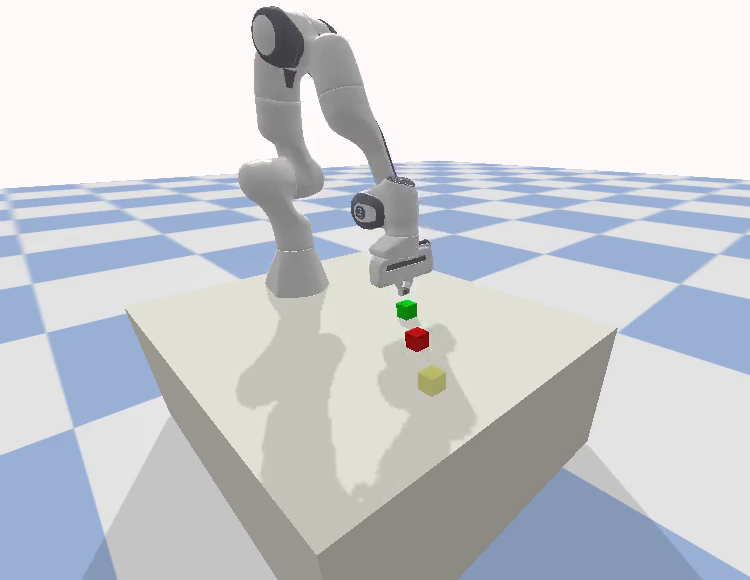}}
          \subcaptionbox*{$\pi_b:t=20$}
        {\includegraphics[width=0.15\linewidth]{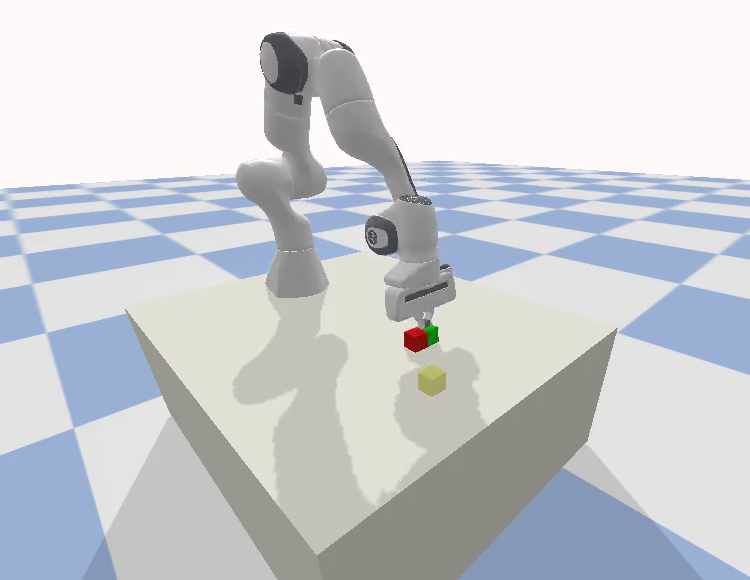}}
        \subcaptionbox*{$\pi_b:t=30$}
        {\includegraphics[width=0.15\linewidth]{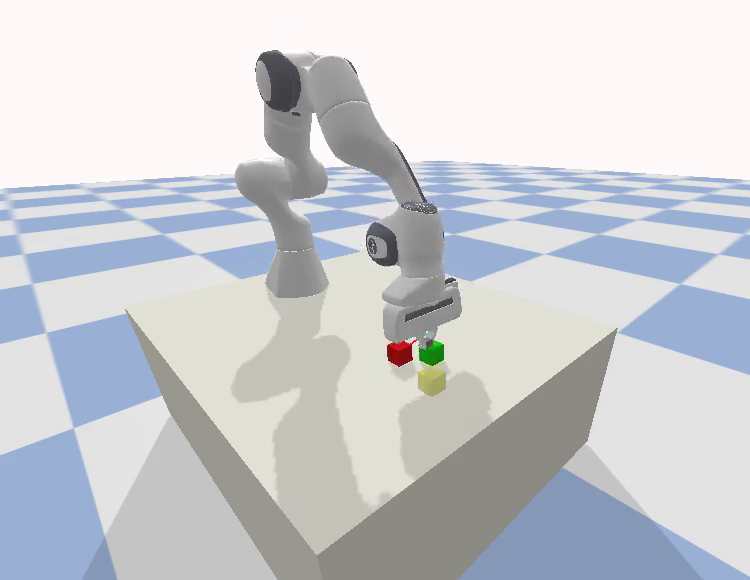}}
              \subcaptionbox*{$\pi_b:t=40$}
        {\includegraphics[width=0.15\linewidth]{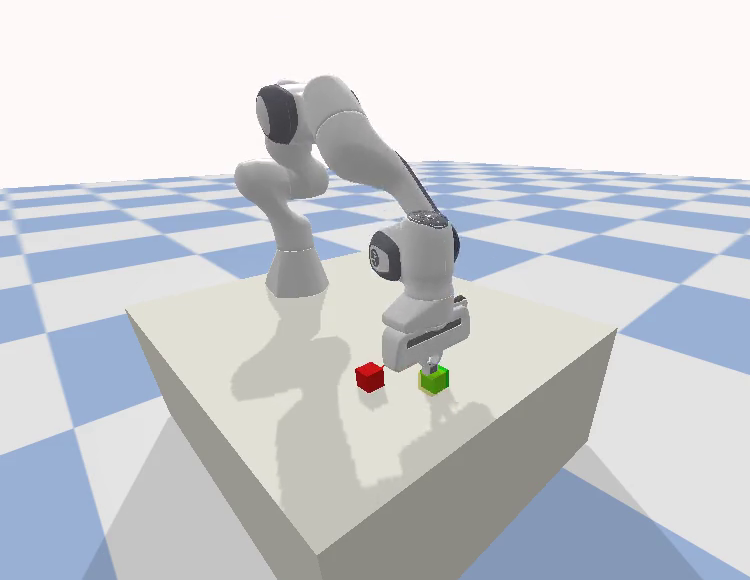}}
        \subcaptionbox*{$\pi_b:t=50$}
        {\includegraphics[width=0.15\linewidth]{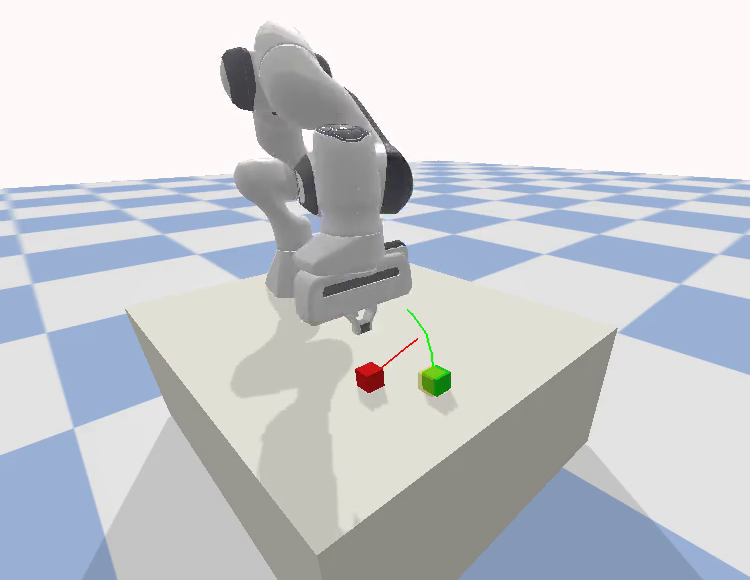}}
      \\  
    \subcaptionbox*{$\pi_s:t=0$}
        {\includegraphics[width=0.15\linewidth]{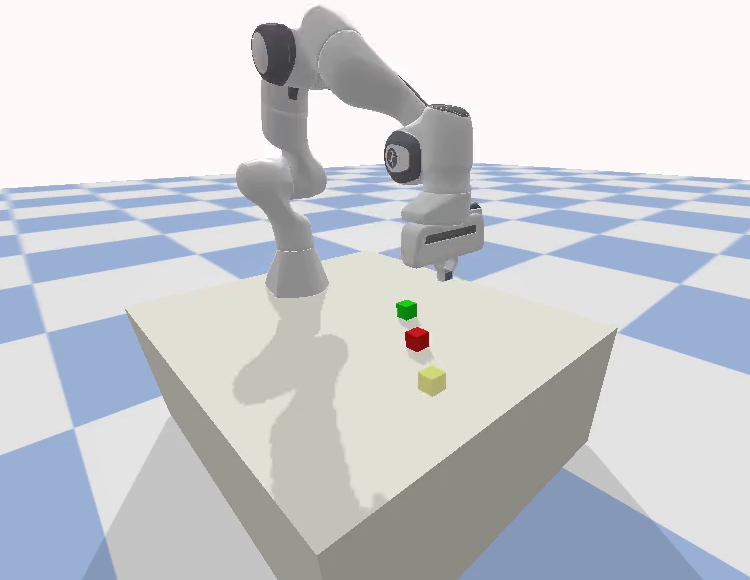}}
        \subcaptionbox*{$\pi_s:t=10$}
        {\includegraphics[width=0.15\linewidth]{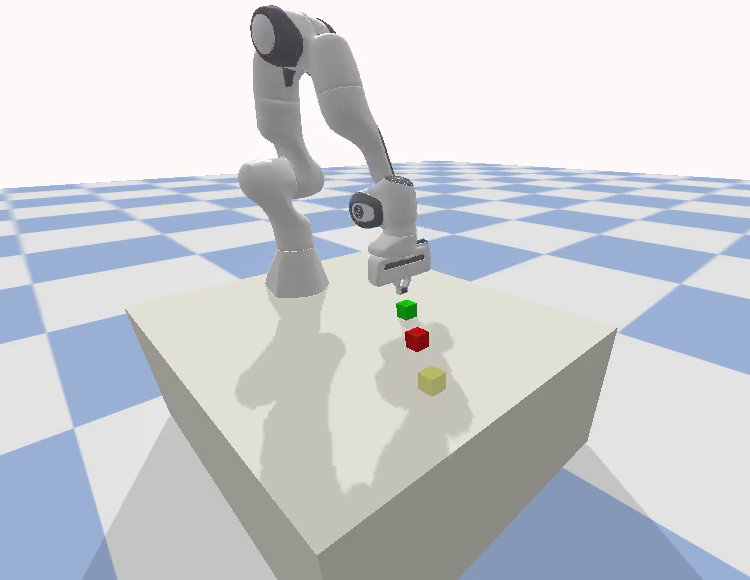}}
          \subcaptionbox*{$\pi_s:t=20$}
        {\includegraphics[width=0.15\linewidth]{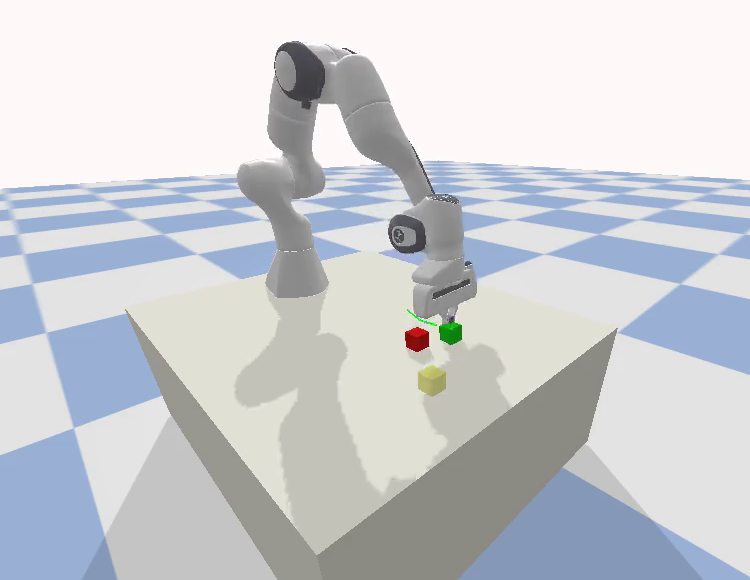}}
        \subcaptionbox*{$\pi_s:t=30$}
        {\includegraphics[width=0.15\linewidth]{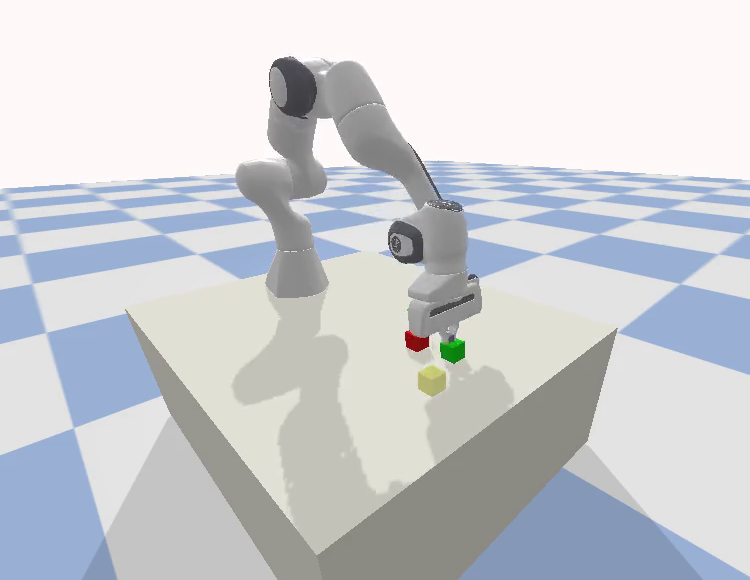}}
              \subcaptionbox*{$\pi_s:t=40$}
        {\includegraphics[width=0.15\linewidth]{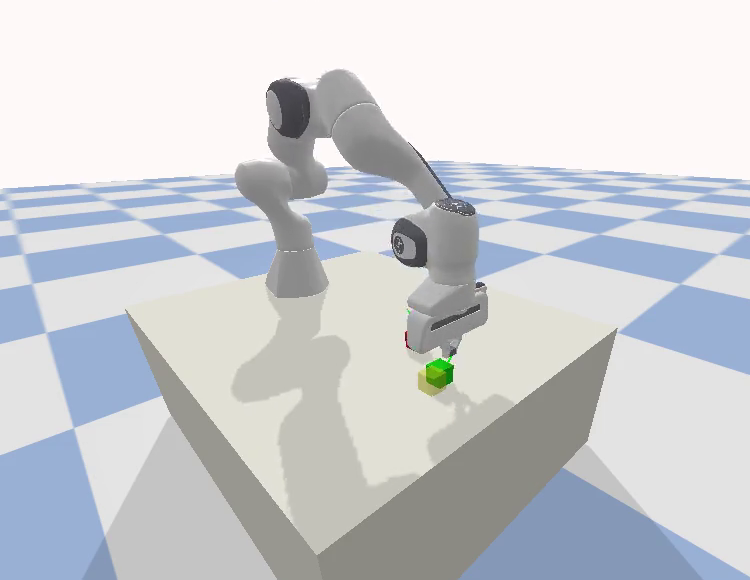}}
        \subcaptionbox*{$\pi_s:t=50$}
        {\includegraphics[width=0.15\linewidth]{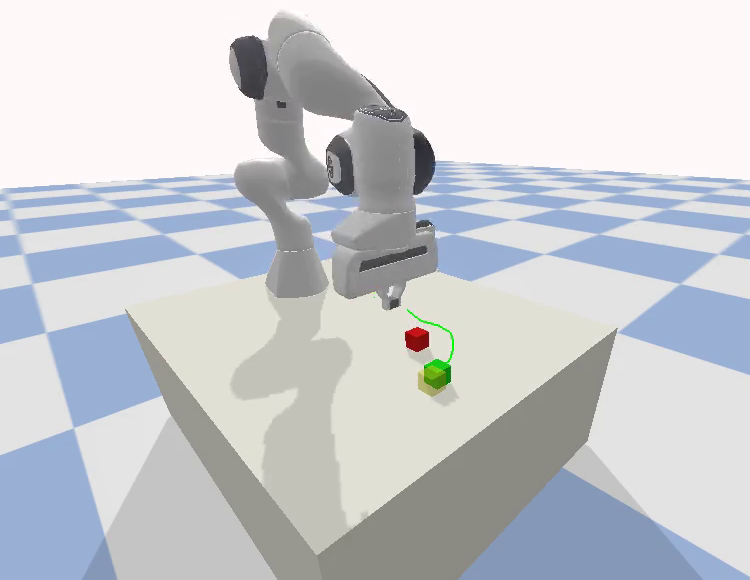}}
      \caption{Robotic arm manipulation task. The agent is rewarded for moving the green cube to the destination in yellow, but violates the safety requirement if it collides with the movable red cube. Rollouts of the baseline policy $\pi_B$ and the safe policy $\pi_S$ are shown on the top line and bottom line, respectively. }
      \label{fig:panda_base}
\end{figure*}
\section{EXPERIMENTS}
% 想回答的问题：
% 1. 我们的算法是否能够从0开始同步训练，两个agent是否能达到各自目标
% 2. 我们的算法是否能从pre-trained unsafe model中直接校正得到
% 3. 相比于sota的Safe RL算法，我们的算法快了多少倍，以及最终的性能相差多少
% 4. 对于Safe RL不能完成的困难任务，我们的解耦算法是否能成功

In the experiments, we answer the following questions:
\begin{enumerate}
\item Can the baseline agent and the target safe agent learn from scratch synchronously and achieve their goals respectively?
\item Can we correct the safe policy from a pre-trained unsafe model with significantly fewer interactions?
\item Compare to the state-of-art Safe RL algorithms, how much sample efficiency our algorithm improves, and how much difference there is in the final performance?
\item Can our decoupled algorithm leverage pre-trained models and succeed for difficult tasks where previous methods barely learn feasible skills without prior knowledge?
\end{enumerate}

\subsection{Experimental Setup}
We demonstrate the proposed algorithm via two challenging safe robot learning tasks, i.e., the quadruped robot locomotion task and the robotic arm manipulation task. In particular, we train the baseline policy and safe policy synchronously from scratch in the quadruped robot locomotion task, whereas we utilize a fixed pre-trained baseline policy in the robotic arm manipulation task.

The baseline policy (if learnable) and the target safe policy are deterministic with $(400,300)$ hidden layers and activated with $\tanh$ function to restrict the output within action limits. 

We compare our algorithm to the state-of-the-art safe RL algorithms, including Constrained Policy Optimization (CPO)~\cite{DBLP:conf/icml/AchiamHTA17}, Lagrangian relaxation with various RL algorithms (i.e. PPO-L, TRPO-L and SAC-L)~\cite{ray2019benchmarking} and the safety layer correction~\cite{DBLP:journals/corr/abs-1801-08757}. To be fair in comparison, we use online supervised learning to initialize agents in the above algorithms if we have already had a baseline controller.

Each experiment is tested over three random seeds.
\subsection{Environment Details}

\subsubsection{Quadruped Robot Locomotion Environment}
As illustrated in Fig.~\ref{fig:ant}, the four-legged ant has to run along the x-direction and is penalized ($I_t = 1$) for exceeding the velocity limit $1.5m/s$ or crossing the safety boundaries. The observation ($\mathcal{S} \in \mathbb{R}^{33}$) includes the information on the position, velocity and quaternion of each link and joint. The action ($\mathcal{A} \in \mathbb{R}^8$) consists of each joint's torque. $\mathcal{R}$ is calculated by x-velocity.
 In this experiment, we set time horizon $T = 500$ and $\delta = 0.05$ (i.e. 0.95-safe policy correction).
\subsubsection{Robotic Arm Manipulation Environment}
The simulated environment is built based on the Panda-gym~\cite{DBLP:journals/corr/abs-2106-13687}, where
the 7-DoF Franka Emika Panda manipulator is used to implement the push task as in Fig.~\ref{fig:panda_base}. The agent has to push the objective (denoted by a green box) to a target position (denoted by a yellow shadow), but is penalized ($I_t = 1$) for colliding with the obstacle (denoted by a red box). We adopt position control to move the manipulator end-effector, i.e., the action ($\mathcal{A} \in \mathbb{R}^3$) is the increments on X-Y-Z axis. The observation ($\mathcal{S} \in \mathbb{R}^{21}$) consists of (1) the velocity and position of the end-effector, (2) the velocity, rotation and position of the objective, (3) the obstacle position and target position. The environment returns a sparse reward (0 for finished and -1 for unfinished). The time horizon is set as $T=50$, and the positions of the objective and target will be randomly initialized at each round. The obstacle will be placed in the way to target.

\subsection{Results on Safe Locomotion Task}
\begin{figure}
      \centering
      \includegraphics[width=0.925\linewidth,trim=0 0 0 10,clip]{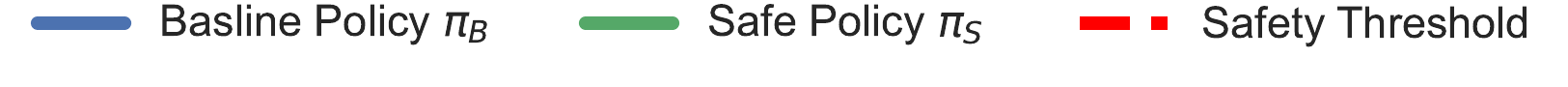}
      \subcaptionbox{Average Episode Return\label{fig:antbase}}
        {\includegraphics[width=0.475\linewidth]{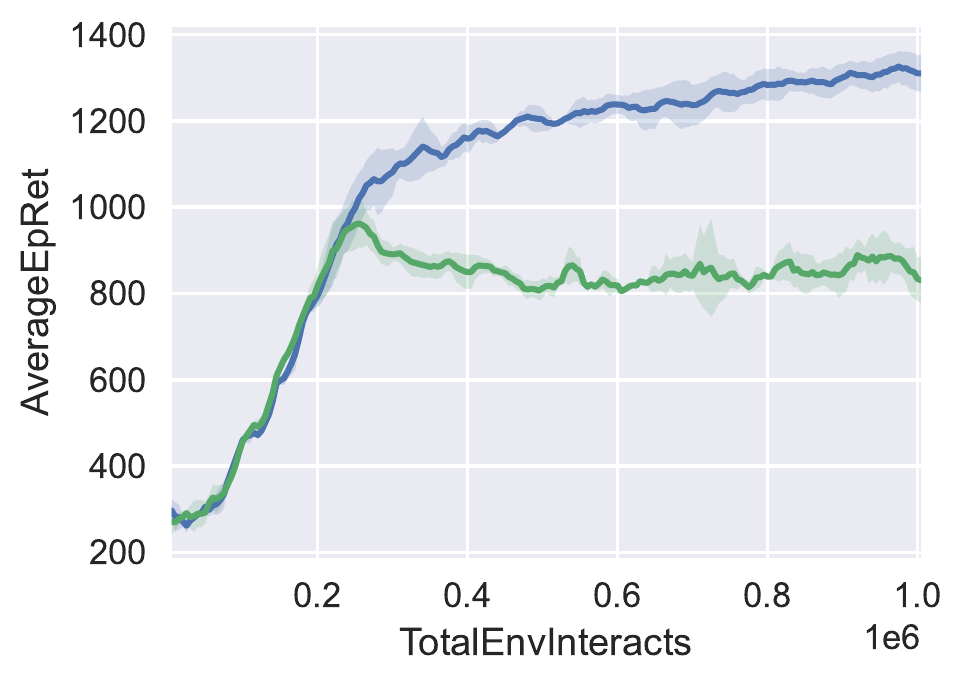}}
      \subcaptionbox{Average Episode Cost\label{fig:antsafe}}
        {\includegraphics[width=0.475\linewidth]{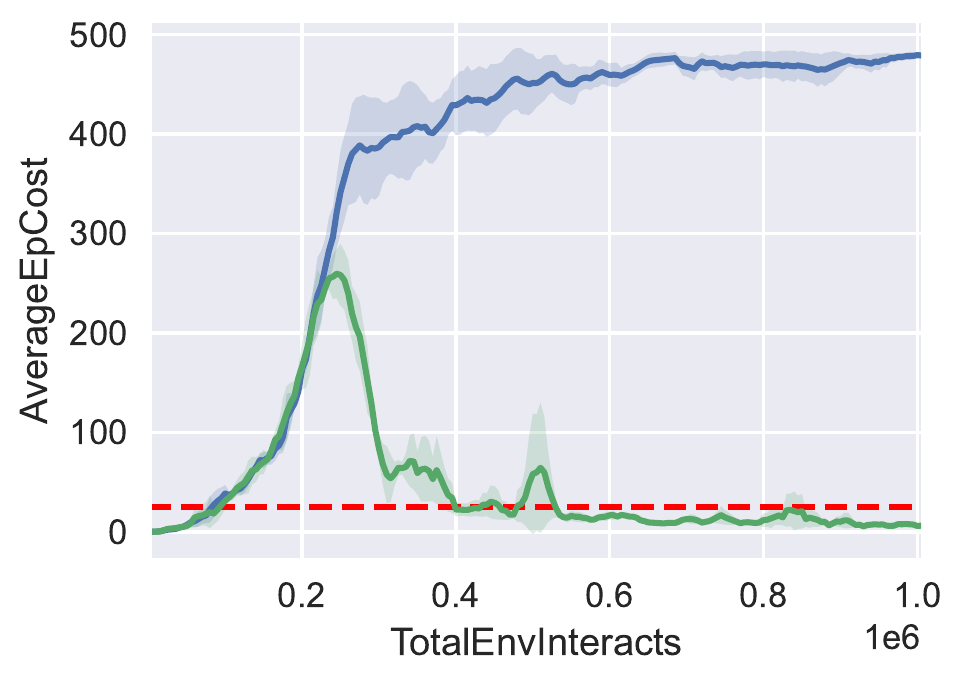}}
      \caption{Learning curves of the baseline policy $\pi_B$ and the safe policy $\pi_S$ on the safe locomotion task. The two agents are trained synchronously in loop. The x-axis is the number of interactions with the emulator. The solid line is the mean and the shaded area is the standard deviation. The dashed line is the expected threshold for 0.95-safe policy.}
      \label{fig:antcurves}
\end{figure}

In this task, the baseline agent and the safe agent are trained synchronously from scratch. Fig.~\ref{fig:antcurves} shows the learning processes of the baseline policy $\pi_B$ and the target safe policy $\pi_S$. At the beginning stage, $\lambda$ is close to zero. Thus $\pi_S$ imitates unconstrained $\pi_B$, and improves its performance at almost the same speed. The safe agent estimates the future cost and becomes risk-aware via past experiences. As  $\lambda$ increases, the penalty for unsafe behaviors dominates the objective function and the return curve of $\pi_S$ in Fig.~\ref{fig:antbase} falls behind $\pi_B$ gradually. Meanwhile, the cost curve of  $\pi_S$ in Fig.~\ref{fig:antsafe} drops quickly and satisfies the 0.95-safe requirement at convergence. By contrast, the average episode cost of baseline policy $\pi_B$ is close to the episode horizon $500$, which means it violates the safety constraint at almost every step. 

As we discussed in Section~\ref{sec:method}, the decoupled formulation often converges to a reasonable and near-optimal policy at a fast speed. In Table~\ref{tab:ant}, we report the performance of our method and other Safe RL methods that solve problem~\eqref{P} directly. We find our approach has much better sample efficiency than baseline algorithms. After $1\times 10^6$ interactions which are relatively expensive in the real-world problem,  our method has already converged to a desirable safe policy, whereas other methods still struggle for reward improvement. Generally, previous Safe RL algorithms need 10 to 20 times experience samples than ours for convergence.

It should be noted that the formulation of safe policy correction won't guarantee a theoretically optimal solution, thus in this task traditional Safe RL algorithms that directly solve~\eqref{P} would achieve sightly better performance at convergence. Nevertheless, the poor data efficiency makes them impractical in use.

\begin{table}

\begin{tabular}{cccccc}
\hline
~ &\multirow{2}*{Ours} & \multirow{2}*{CPO} & \multirow{2}*{PPO-L} & \multirow{2}*{TRPO-L} & \multirow{2}*{SAC-L}  \\
~ & ~ & ~ & ~ & ~ & ~ \\
\hline
Return & \textbf{898.20} & 550.43 & 571.23 & 546.45 & 329.87 \\
\scriptsize{(1E6 steps)} & $\pm12.29$ & $\pm8.86$ & $\pm11.03$ & $\pm12.55$ & $\pm27.65$ \\
%\hline
Cost & \textbf{1.77} & 3.63 & 3.33 & 3.49 & 2.88 \\
\scriptsize{(1E6 steps)} & $\pm0.22$ & $\pm0.86$ & $\pm0.83$ & $\pm0.54$ & $\pm0.26$ \\
%\hline
Return & 898.20 & 915.36 & \textbf{944.27} & 917.63 & 450.24 \\
\scriptsize{(Convergence)} & $\pm12.29$ & $\pm10.23$ & $\pm12.77$ & $\pm11.27$ & $\pm22.59$ \\
%\hline
Cost & \textbf{1.77} & 2.65 & 2.71 & 2.15 & 2.12 \\
\scriptsize{(Convergence)}  & $\pm0.22$ & $\pm0.49$ & $\pm0.51$ & $\pm0.57$ & $\pm0.20$ \\
%\hline
\multirow{2}*{
Total samples} & \textbf{5E5} & 1E7 & 1E7 & 1E7 & 6E6 \\
~ & 1x & 20x & 20x & 20x & 12x \\
\hline
\end{tabular}
\caption{We report average episode return (cumulative rewards) and cost (safety violations) on the safe locomotion task after $10^6$ interactions and at convergence. We also show the sample multiples required by baseline algorithms compared to our method.}
\label{tab:ant}
\end{table}

\subsection{Results on Safe Manipulation Task}

In this task, the baseline agent is pre-trained with Hindsight Experience Replay (HER)~\cite{andrychowicz2017hindsight}. This method is proved to be effective for standard robotic arm manipulation but hasn't been introduced into Safe RL yet. Thus, it is an actual example explaining how we can directly leverage well-studied techniques in standard RL to our decoupled safe robot learning framework.

\begin{figure}
      \centering
      \includegraphics[width=0.925\linewidth,trim=0 0 0 10,clip]{figures/_legend.pdf}
      \subcaptionbox{Average Episode Return\label{fig:pandabase}}
        {\includegraphics[width=0.475\linewidth]{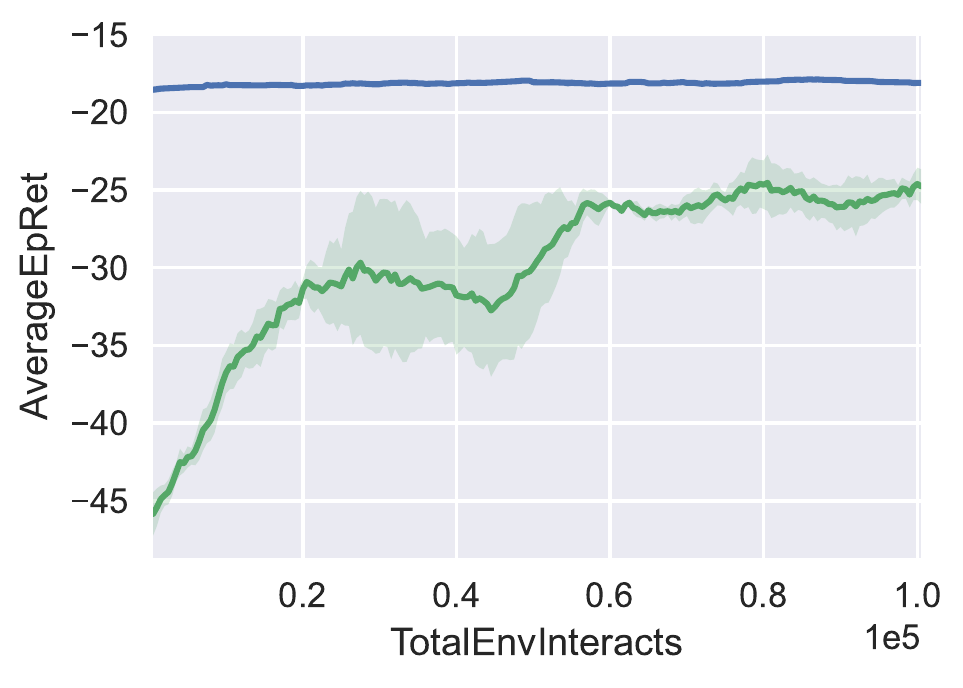}}
      \subcaptionbox{Average Episode Cost\label{fig:pandasafe}}
        {\includegraphics[width=0.475\linewidth]{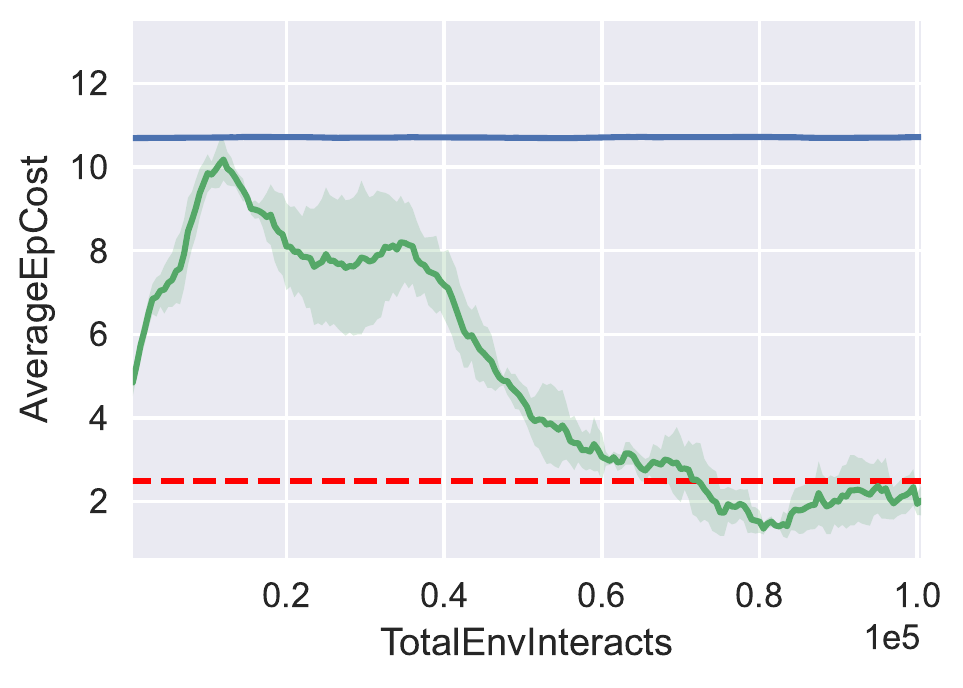}}
      \caption{Learning curves of the safe policy $\pi_S$ on the safe manipulation task. The pre-trained baseline policy $\pi_B$ is fixed during the process. The x-axis is the number of interactions with the emulator. The solid line is the mean and the shaded area is the standard deviation. The dashed line is the expected threshold for 0.95-safe policy.}
      \label{fig:pandacurves}
\end{figure}

Fig.~\ref{fig:pandacurves} shows the learning processes of the safe policy $\pi_S$. At the very beginning, the agent learns some helpful but unsafe behaviors from the fixed baseline. Nevertheless, the off-policy safe RL tuning prohibits it from acting the same as the baseline but modifies its outputs to reduce safety violations under the expected threshold, as illustrated in Fig.~\ref{fig:pandasafe}. Also, moving the cube on a curved trajectory degrades the maximum of episode return.

\begin{figure}
      \centering
      \includegraphics[width=0.925\linewidth,trim=0 0 0 10,clip]{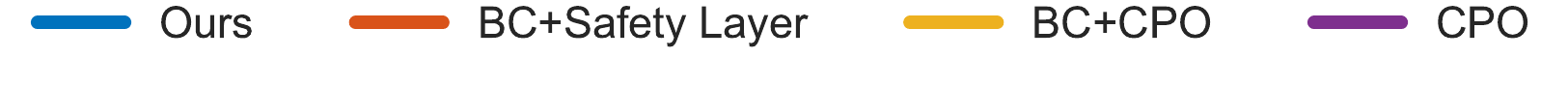}
      \includegraphics[width=0.925\linewidth,trim=175 120 50 150,clip]{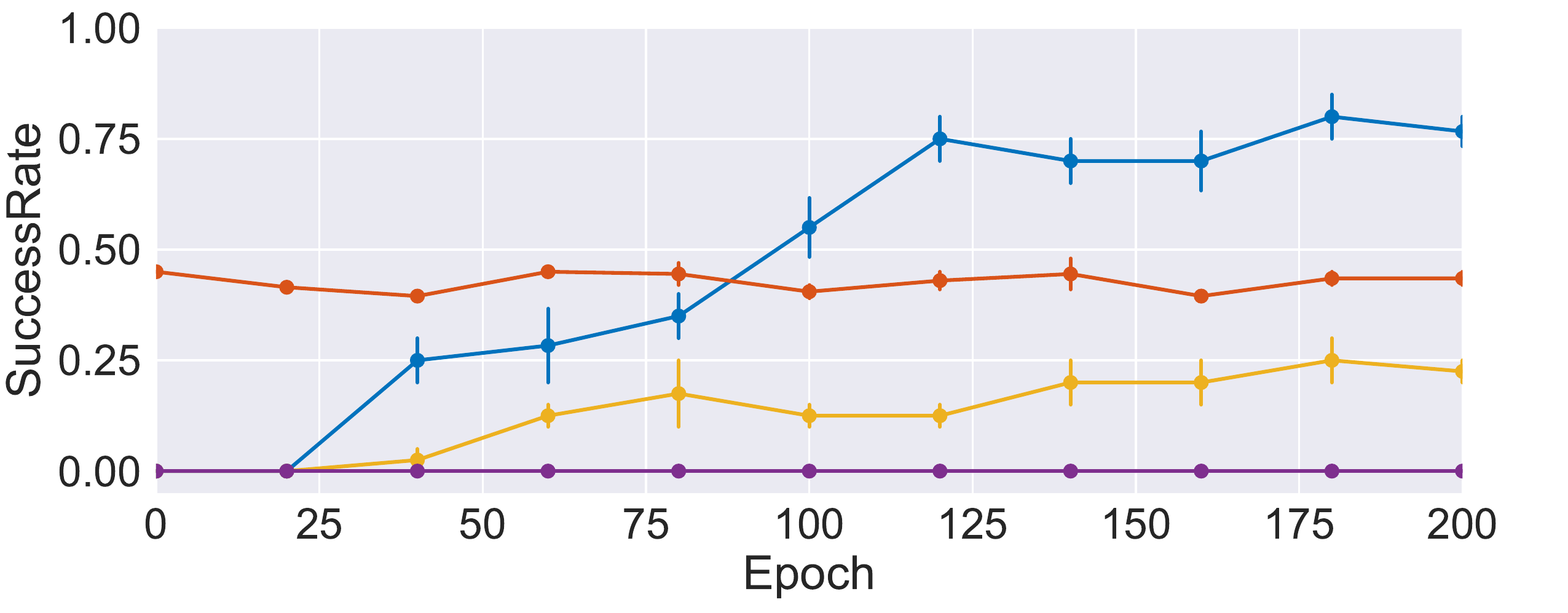}
      \caption{Success rate of different algorithms on the safe manipulation task. A success round counts if the target object is pushed to the destination without the obstacle touched. 20 rounds are evaluated and averaged every 25 epoch (500 steps for each epoch).}
      \label{fig:pandacurves2}
\end{figure}

We also plot the evaluation curves of success rate for different algorithms on this task in Fig.~\ref{fig:pandacurves2}. Notably, the cube-push task is too difficult to learn for any previous Safe RL algorithm, and none of them can successfully finish the job even once time. Here, we only take CPO as a representative in the plot. In contrast, our approach corrects the safe policy from the pre-trained agent, which requires a small number of interactions and has a high success rate at around 80\% finally. To leverage the pre-trained model and be fair in comparison, we also perform the behavior cloning to initialize the agent in Safety Layer and CPO algorithms. This helps them obtain basic skills, but it is still hard to entirely avoid the risk. The success rate stays below 30\% after the same number of interactions as our method. We also incorporate a safety layer into the pre-trained model via offline data. The first-order Taylor's approximation on risk-estimation and overly short-sighted safe action correction limit its success rate at around 50\%. Conclusively, our approach has the best 
the trade-off for desirable performance and sample efficiency. 

\section{CONCLUSIONS}

In this paper, we propose a dual-agent risk-aware policy learning algorithm for safe policy in robotics. The main idea is to decouple the task into the baseline policy learning and the safe policy correction. The baseline agent can leverage useful techniques in standard RL and even non-learning-based controllers in typical robotic applications; The safe agent is corrected from the baseline with limited data via online behavioral cloning and off-policy constrained RL tuning. Compared to previous Safe RL algorithms, our approach is more data efficient for sample-expensive robotic tasks and achieves more extensive exploration for mastering hard-to-learn skills. Experimental results demonstrate that the proposed method is effective on different, challenging safe robot learning tasks and can obtain a safe and reasonable solution much faster than  prior work.

%\addtolength{\textheight}{-12cm}   % This command serves to balance the column lengths
                                  % on the last page of the document manually. It shortens
                                  % the textheight of the last page by a suitable amount.
                                  % This command does not take effect until the next page
                                  % so it should come on the page before the last. Make
                                  % sure that you do not shorten the textheight too much.

%%%%%%%%%%%%%%%%%%%%%%%%%%%%%%%%%%%%%%%%%%%%%%%%%%%%%%%%%%%%%%%%%%%%%%%%%%%%%%%%

%%%%%%%%%%%%%%%%%%%%%%%%%%%%%%%%%%%%%%%%%%%%%%%%%%%%%%%%%%%%%%%%%%%%%%%%%%%%%%%%

%%%%%%%%%%%%%%%%%%%%%%%%%%%%%%%%%%%%%%%%%%%%%%%%%%%%%%%%%%%%%%%%%%%%%%%%%%%%%%%%
%\section*{APPENDIX}

%\section*{ACKNOWLEDGMENT}

%\clearpage
\bibliographystyle{unsrt}       % Include this if you use bibtex
\bibliography{root}  

\begin{thebibliography}{10}

\bibitem{francis2020long}
Anthony Francis, Aleksandra Faust, Hao-Tien~Lewis Chiang, Jasmine Hsu, J~Chase
  Kew, Marek Fiser, and Tsang-Wei~Edward Lee.
\newblock Long-range indoor navigation with prm-rl.
\newblock {\em IEEE Transactions on Robotics}, 36(4):1115--1134, 2020.

\bibitem{rajeswaran2017learning}
Aravind Rajeswaran, Vikash Kumar, Abhishek Gupta, Giulia Vezzani, John
  Schulman, Emanuel Todorov, and Sergey Levine.
\newblock Learning complex dexterous manipulation with deep reinforcement
  learning and demonstrations.
\newblock {\em arXiv preprint arXiv:1709.10087}, 2017.

\bibitem{shin2019obstacle}
Sang-Yun Shin, Yong-Won Kang, and Yong-Guk Kim.
\newblock Obstacle avoidance drone by deep reinforcement learning and its
  racing with human pilot.
\newblock {\em Applied sciences}, 9(24):5571, 2019.

\bibitem{amodei2016concrete}
Dario Amodei, Chris Olah, Jacob Steinhardt, Paul Christiano, John Schulman, and
  Dan Man{\'e}.
\newblock Concrete problems in ai safety.
\newblock {\em arXiv preprint arXiv:1606.06565}, 2016.

\bibitem{hewing2020learning}
Lukas Hewing, Kim~P Wabersich, Marcel Menner, and Melanie~N Zeilinger.
\newblock Learning-based model predictive control: Toward safe learning in
  control.
\newblock {\em Annual Review of Control, Robotics, and Autonomous Systems},
  3:269--296, 2020.

\bibitem{sui2015safe}
Yanan Sui, Alkis Gotovos, Joel Burdick, and Andreas Krause.
\newblock Safe exploration for optimization with gaussian processes.
\newblock In {\em International conference on machine learning}, pages
  997--1005. PMLR, 2015.

\bibitem{brunke2021safe}
Lukas Brunke, Melissa Greeff, Adam~W Hall, Zhaocong Yuan, Siqi Zhou, Jacopo
  Panerati, and Angela~P Schoellig.
\newblock Safe learning in robotics: From learning-based control to safe
  reinforcement learning.
\newblock {\em Annual Review of Control, Robotics, and Autonomous Systems}, 5,
  2021.

\bibitem{ray2019benchmarking}
Alex Ray, Joshua Achiam, and Dario Amodei.
\newblock Benchmarking safe exploration in deep reinforcement learning.
\newblock {\em arXiv preprint arXiv:1910.01708}, 7, 2019.

\bibitem{DBLP:conf/icml/AchiamHTA17}
Joshua Achiam, David Held, Aviv Tamar, and Pieter Abbeel.
\newblock Constrained policy optimization.
\newblock In Doina Precup and Yee~Whye Teh, editors, {\em Proceedings of the
  34th International Conference on Machine Learning, {ICML} 2017, Sydney, NSW,
  Australia, 6-11 August 2017}, volume~70 of {\em Proceedings of Machine
  Learning Research}, pages 22--31. {PMLR}, 2017.

\bibitem{andrychowicz2017hindsight}
Marcin Andrychowicz, Filip Wolski, Alex Ray, Jonas Schneider, Rachel Fong,
  Peter Welinder, Bob McGrew, Josh Tobin, OpenAI Pieter~Abbeel, and Wojciech
  Zaremba.
\newblock Hindsight experience replay.
\newblock {\em Advances in neural information processing systems}, 30, 2017.

\bibitem{DBLP:conf/ijcai/LiuHL21}
Yongshuai Liu, Avishai Halev, and Xin Liu.
\newblock Policy learning with constraints in model-free reinforcement
  learning: {A} survey.
\newblock In Zhi{-}Hua Zhou, editor, {\em Proceedings of the Thirtieth
  International Joint Conference on Artificial Intelligence, {IJCAI} 2021,
  Virtual Event / Montreal, Canada, 19-27 August 2021}, pages 4508--4515.
  ijcai.org, 2021.

\bibitem{9029423}
Santiago Paternain, Miguel Calvo-Fullana, Luiz F.~O. Chamon, and Alejandro
  Ribeiro.
\newblock Learning safe policies via primal-dual methods.
\newblock In {\em 2019 IEEE 58th Conference on Decision and Control (CDC)},
  pages 6491--6497, 2019.

\bibitem{DBLP:journals/corr/abs-1801-08757}
Gal Dalal, Krishnamurthy Dvijotham, Matej Vecer{\'{\i}}k, Todd Hester, Cosmin
  Paduraru, and Yuval Tassa.
\newblock Safe exploration in continuous action spaces.
\newblock {\em CoRR}, abs/1801.08757, 2018.

\bibitem{DBLP:conf/iros/HirshbergVK20}
Tom Hirshberg, Sai Vemprala, and Ashish Kapoor.
\newblock Safety considerations in deep control policies with safety barrier
  certificates under uncertainty.
\newblock In {\em {IEEE/RSJ} International Conference on Intelligent Robots and
  Systems, {IROS} 2020, Las Vegas, NV, USA, October 24, 2020 - January 24,
  2021}, pages 6245--6251. {IEEE}, 2020.

\bibitem{DBLP:conf/aaai/HesterVPLSPHQSO18}
Todd Hester, Matej Vecer{\'{\i}}k, Olivier Pietquin, Marc Lanctot, Tom Schaul,
  Bilal Piot, Dan Horgan, John Quan, Andrew Sendonaris, Ian Osband, Gabriel
  Dulac{-}Arnold, John~P. Agapiou, Joel~Z. Leibo, and Audrunas Gruslys.
\newblock Deep q-learning from demonstrations.
\newblock In Sheila~A. McIlraith and Kilian~Q. Weinberger, editors, {\em
  Proceedings of the Thirty-Second {AAAI} Conference on Artificial
  Intelligence, (AAAI-18), the 30th innovative Applications of Artificial
  Intelligence (IAAI-18), and the 8th {AAAI} Symposium on Educational Advances
  in Artificial Intelligence (EAAI-18), New Orleans, Louisiana, USA, February
  2-7, 2018}, pages 3223--3230. {AAAI} Press, 2018.

\bibitem{DBLP:journals/corr/VecerikHSWPPHRL17}
Matej Vecer{\'{\i}}k, Todd Hester, Jonathan Scholz, Fumin Wang, Olivier
  Pietquin, Bilal Piot, Nicolas Heess, Thomas Roth{\"{o}}rl, Thomas Lampe, and
  Martin~A. Riedmiller.
\newblock Leveraging demonstrations for deep reinforcement learning on robotics
  problems with sparse rewards.
\newblock {\em CoRR}, abs/1707.08817, 2017.

\bibitem{DBLP:conf/icml/KangJF18}
Bingyi Kang, Zequn Jie, and Jiashi Feng.
\newblock Policy optimization with demonstrations.
\newblock In Jennifer~G. Dy and Andreas Krause, editors, {\em Proceedings of
  the 35th International Conference on Machine Learning, {ICML} 2018,
  Stockholmsm{\"{a}}ssan, Stockholm, Sweden, July 10-15, 2018}, volume~80 of
  {\em Proceedings of Machine Learning Research}, pages 2474--2483. {PMLR},
  2018.

\bibitem{DBLP:conf/aaai/JingMHSY0020}
Mingxuan Jing, Xiaojian Ma, Wenbing Huang, Fuchun Sun, Chao Yang, Bin Fang, and
  Huaping Liu.
\newblock Reinforcement learning from imperfect demonstrations under soft
  expert guidance.
\newblock In {\em The Thirty-Fourth {AAAI} Conference on Artificial
  Intelligence, {AAAI} 2020, The Thirty-Second Innovative Applications of
  Artificial Intelligence Conference, {IAAI} 2020, The Tenth {AAAI} Symposium
  on Educational Advances in Artificial Intelligence, {EAAI} 2020, New York,
  NY, USA, February 7-12, 2020}, pages 5109--5116. {AAAI} Press, 2020.

\bibitem{DBLP:conf/iclr/GaoXLYLD18}
Yang Gao, Huazhe Xu, Ji~Lin, Fisher Yu, Sergey Levine, and Trevor Darrell.
\newblock Reinforcement learning from imperfect demonstrations.
\newblock In {\em 6th International Conference on Learning Representations,
  {ICLR} 2018, Vancouver, BC, Canada, April 30 - May 3, 2018, Workshop Track
  Proceedings}. OpenReview.net, 2018.

\bibitem{DBLP:conf/atal/GoecksGLVW20}
Vinicius~G. Goecks, Gregory~M. Gremillion, Vernon~J. Lawhern, John Valasek, and
  Nicholas~R. Waytowich.
\newblock Integrating behavior cloning and reinforcement learning for improved
  performance in dense and sparse reward environments.
\newblock In Amal El~Fallah Seghrouchni, Gita Sukthankar, Bo~An, and Neil
  Yorke{-}Smith, editors, {\em Proceedings of the 19th International Conference
  on Autonomous Agents and Multiagent Systems, {AAMAS} '20, Auckland, New
  Zealand, May 9-13, 2020}, pages 465--473. International Foundation for
  Autonomous Agents and Multiagent Systems, 2020.

\bibitem{DBLP:conf/ijcai/Wang0ZCZ21}
Zhaorong Wang, Meng Wang, Jingqi Zhang, Yingfeng Chen, and Chongjie Zhang.
\newblock Reward-constrained behavior cloning.
\newblock In Zhi{-}Hua Zhou, editor, {\em Proceedings of the Thirtieth
  International Joint Conference on Artificial Intelligence, {IJCAI} 2021,
  Virtual Event / Montreal, Canada, 19-27 August 2021}, pages 3169--3175.
  ijcai.org, 2021.

\bibitem{schafer2021decoupling}
Lukas Sch{\"a}fer, Filippos Christianos, Josiah Hanna, and Stefano~V Albrecht.
\newblock Decoupling exploration and exploitation in reinforcement learning.
\newblock {\em arXiv preprint arXiv:2107.08966}, 2021.

\bibitem{whitney2021decoupled}
William~F Whitney, Michael Bloesch, Jost~Tobias Springenberg, Abbas
  Abdolmaleki, Kyunghyun Cho, and Martin Riedmiller.
\newblock Decoupled exploration and exploitation policies for sample-efficient
  reinforcement learning.
\newblock {\em arXiv preprint arXiv:2101.09458}, 2021.

\bibitem{altman1999constrained}
Eitan Altman.
\newblock {\em Constrained Markov decision processes}, volume~7.
\newblock CRC Press, 1999.

\bibitem{silver2014deterministic}
David Silver, Guy Lever, Nicolas Heess, Thomas Degris, Daan Wierstra, and
  Martin Riedmiller.
\newblock Deterministic policy gradient algorithms.
\newblock In {\em International conference on machine learning}, pages
  387--395. PMLR, 2014.

\bibitem{luenberger1984linear}
David~G Luenberger, Yinyu Ye, et~al.
\newblock {\em Linear and nonlinear programming}, volume~2.
\newblock Springer, 1984.

\bibitem{tessler2018reward}
Chen Tessler, Daniel~J Mankowitz, and Shie Mannor.
\newblock Reward constrained policy optimization.
\newblock In {\em International Conference on Learning Representations}, 2018.

\bibitem{yang2021exploration}
Tianpei Yang, Hongyao Tang, Chenjia Bai, Jinyi Liu, Jianye Hao, Zhaopeng Meng,
  and Peng Liu.
\newblock Exploration in deep reinforcement learning: a comprehensive survey.
\newblock {\em arXiv preprint arXiv:2109.06668}, 2021.

\bibitem{fujimoto2018addressing}
Scott Fujimoto, Herke Hoof, and David Meger.
\newblock Addressing function approximation error in actor-critic methods.
\newblock In {\em International conference on machine learning}, pages
  1587--1596. PMLR, 2018.

\bibitem{chow2019lyapunov}
Yinlam Chow, Ofir Nachum, Aleksandra Faust, Edgar Duenez-Guzman, and Mohammad
  Ghavamzadeh.
\newblock Lyapunov-based safe policy optimization for continuous control.
\newblock {\em arXiv preprint arXiv:1901.10031}, 2019.

\bibitem{DBLP:journals/corr/abs-2106-13687}
Quentin Gallou{\'{e}}dec, Nicolas Cazin, Emmanuel Dellandr{\'{e}}a, and Liming
  Chen.
\newblock Multi-goal reinforcement learning environments for simulated franka
  emika panda robot.
\newblock {\em CoRR}, abs/2106.13687, 2021.

\end{thebibliography}

\end{document}